\documentclass[runningheads]{llncs}
\usepackage{gensymb}
\usepackage{perpage} 
\MakePerPage{footnote} 

\usepackage[T1]{fontenc}
\usepackage{graphicx}
\begin{document}
\title{A Multimodal Haptic Interface for Simultaneous Softness and Thermal Modulation}
\titlerunning{Soft Thermotactile Display}
%
\author{Joshua Brown\inst{1,2} \and
Fernando Bello\inst{1,3} \and
Lynette Jones\inst{2}}
\authorrunning{J. Brown et al.}
%
\institute{Department of Surgery and Cancer, Imperial College London, UK \and
Department of Mechanical Engineering, Massachusetts Institute of Technology, Cambridge, MA, USA \and Duke-NUS Medical School, Singapore\\
\email{joshua.brown@imperial.ac.uk}\\}
\maketitle              
\begin{abstract}
Haptic interactions, such as touching an object in virtual reality, involve detecting a plethora of sensations through the skin. Whilst tactile displays exist to simulate and manipulate all of these cues individually, little progress has been made constructing devices that can control multiple cues simultaneously. This paper presents a tactile display technology that offers both softness and temperature change.

A prototype display, based on particle jamming and the thermal transfer of heated and cooled water, is evaluated in terms of its ability to change both stiffness and temperature. Mechanical evaluations demonstrate controllable stiffness increase of 300\%, whilst thermal characterisation showed strong surface uniformity and a range of achievable temperatures from 10\degree~C up to 50\degree~C, with closed-loop control maintaining a ±0.2\degree~C error from a setpoint. This work advances the design of multimodal tactile displays with applications to perceptual research, medical simulation, and affective haptics.

\keywords{Soft haptics  \and Thermal haptics \and Tactile displays}
\end{abstract}
%
%
\section{Introduction} 

Softness-changing surfaces are becoming increasingly important for creating realistic haptic interactions in VR and medical simulation. Rapid technical advances in this area have been enabled by soft robotics techniques. Softness changing haptic devices have wide-ranging applications, from wearable haptic suits for enhancing entertainment experiences \cite{Kang2023PneumaticSimulation,AlMaimani2017FrozenGames}, to simulation-based training for medical procedures and examinations with realistic soft tissues \cite{Brown2024HardnessTissues}. Softness may be approximated via impedance control of haptic robots, emulating force-deformation properties of compliant materials \cite{Mohsen2003HapticBody}, by directly altering the properties of a surface or device to change its compliance \cite{Stanley2013}, or a combination of the two \cite{Mete2024SORI:Perception}.

Particle jamming is an increasingly popular technology for creating tactile displays that can alter their softness. In a jamming system, granular media, which are normally highly malleable, may be stiffened by compressing their volume, causing the individual particles to brace against each other, or `jam', thereby creating the appearance of a solid mass \cite{vanHecke2010JammingIsostaticity}. Jamming has been applied to haptic surfaces that can change their softness to influence user interaction \cite{Follmer2012}, retain their shape after being moulded by a user \cite{Sato2014}, or simulate palpation in medical training \cite{Li2014Multi-FingeredActuators}. It has been applied to gloves \cite{Simon2014WearableHaptics} and exosuits \cite{AlMaimani2017FrozenGames}, to restrict movement by stiffening the joints in these wearable devices. Particle jamming has also been proposed as a mechanism for creating `remote touch' - perceiving solid objects buried in granular media \cite{Chen2025ExploringStudy}, as well as a basis for multimodal cutaneous interfaces that can leverage the adaptability of jamming technology as a medium for generating other cutaneous sensations \cite{Brown2025ROVHaptics}. This has become possible with the advent of fully mechanical jamming technologies that do not rely on airtight containers or pneumatic systems \cite{Brown2024HardnessTissues}.

Whilst softness has many important applications in haptics, it is just one of several cutaneous sensations that enable humans to perceive the world via touch \cite{Jones2018HowTouch}. Thermal sensations resulting from changes in skin temperature also play an important and varied role, from warning of danger due to the skin burning or freezing, to playing a more nuanced role in multisensory cutaneous experiences, such as perceiving wetness \cite{BergmannTiest2012HapticWetness}, or discriminating between physical materials based on the rate with which skin temperature changes \cite{Ho2018MaterialApplications}.  Such cues can even influence the `pleasantness' of different tactile experiences \cite{Ackerley2014HumanCaress}. Thermal sensations have also been shown to influence the other senses via crossmodal correspondences, such as the associations between colour and temperature (red-hot, blue-cold) \cite{Spence2020Temperature-BasedConsequences}. Thermal cues can be generated in a number of ways, including directly using thermoelectric Peltier modules \cite{Sines1999PeltierEnvironments}, via convection using heated and cooled air \cite{Singhal2021Mid-AirDisplay} or water \cite{Liu2021ThermoCaress:Stimulation}, or through radiation via infrared light \cite{LecuyerHOMERE:Environments}.

With these applications in mind, there is a strong motivation for creating tactile display technology capable of producing both softness and thermal sensations simultaneously. Such a device could be used to simulate thermal properties of vascular disease in medical training, or the varying stiffness and thermal emissivity of different materials in a VR simulation. This paper presents a novel multimodal surface that integrates constrictive particle jamming with a fluidic thermal skin. Key contributions are: a demonstration of how pump-free jamming can mediate otherwise incompatible technologies; a mechanical design that optimizes thermal transfer without compromising jamming compliance; and an experimental characterisation of the device's performance.

\section{Hardware design}

\subsection{Softness}
Softness change is achieved using particle jamming, and specifically the constrictive approach described in \cite{Brown2024HardnessTissues}, whereby particles are held in a fabric container which is twisted from one end, exerting radial forces on the granular fluid in all directions, causing it to jam. The implementation for the prototype described here departed from the earlier design in that it used a circular shape to improve the uniformity of the jamming effect (which was previously reported as being harder in the centre than the corners), as well as being larger, with a 150~mm diameter and 30~mm depth. It is constructed in a 3D printed plastic frame produced in PLA plastic using the `0.2~mm strength' profile on a Bambu Lab P1S 3D printer. To account for the larger volume of particles, a more powerful Maxon EC90 brushless DC motor with a nominal torque output of 1.26~Nm was selected as an actuator. A Maxon EPOS4 70/15 controller was used to provide closed-loop torque control, enabling faster and more accurate control of the display's stiffness.

\subsection{Temperature}
Thermal feedback is provided by a thin silicone skin attached to the top surface of the jamming container. Channels to route the liquid through the skin are cast from Smooth-On Dragon Skin 10 FAST, and attached to a touch surface, made from a composite of Dragon Skin 10 FAST and copper powder (mesh size 325~$\mu$m) at a 60:40 ratio by mass. The copper composite layer provides a thermally conductive interface between adjacent channels, as well as the temperature controlled liquid and a user's finger. The pure silicone insulates the large thermal mass of the particle fluid.

\begin{figure}[h!]
\centering
\includegraphics[width=\textwidth]{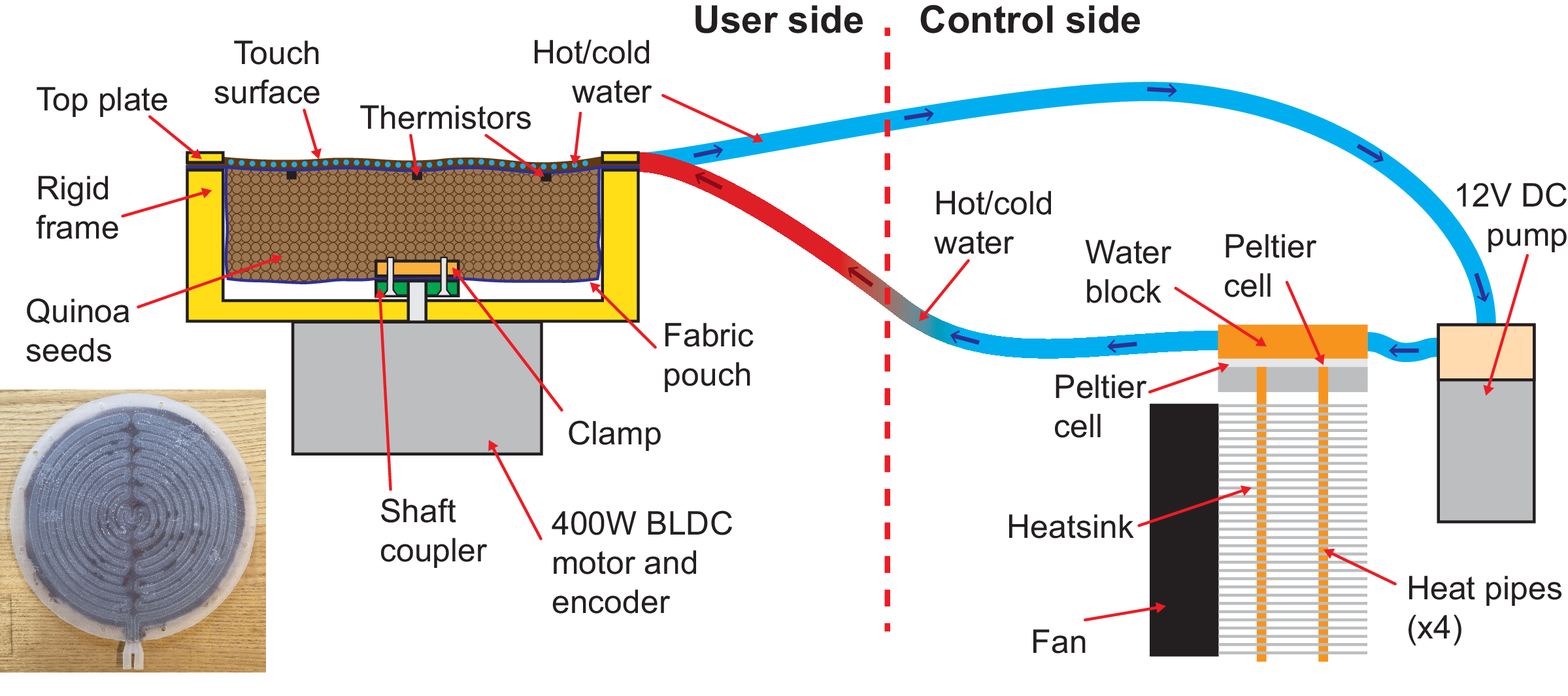}
\caption{Schematic view of the mechanical and hydraulic components of the combined softness and temperature changing tactile display. Insert: The underside of the cast silicone skin with fluid channels visible.} \label{fig:schematic}
\end{figure}

The liquid temperature is controlled by a custom control unit based on a 100~W  Peltier module (Model ET-161-12-08-E, European Thermal Dynamics, UK) with a 40~mm $\times$ 40~mm copper water block used to transfer heat into and out of the water in a closed-loop design. A Cooler Master Hyper 212 EVO CPU heatsink and fan were attached to the other side of the Peltier module to remove excess heat during cooling. Thermal adhesive was used to attach all three parts of this stack. An insulated cover was placed over the water block to limit thermal interference from the environment. The  system is controlled by an Arduino Giga R1 Wifi microcontroller running a hybrid bang-bang/PID control-loop based on feedback from Amphenol 10~k$\Omega$ thermistors placed on the bottom of the thermal skin. In this arrangement, bang-bang control is used when the difference between the display temperature and setpoint is greater than ±5\degree~C, and PID control is used inside this range. This reduces integration error due to the relatively slow thermal response. The system schematic is shown in Figure \ref{fig:schematic}, with the hardware shown in Figure \ref{fig:setup} right.

\section{Performance and characterisation}
The performance of the device was characterised in terms of both the range of its achievable softness change, and the range and consistency of its achievable surface temperature.

\subsection{Softness}

\begin{figure}[h!]
\centering
\includegraphics[width=0.95\textwidth]{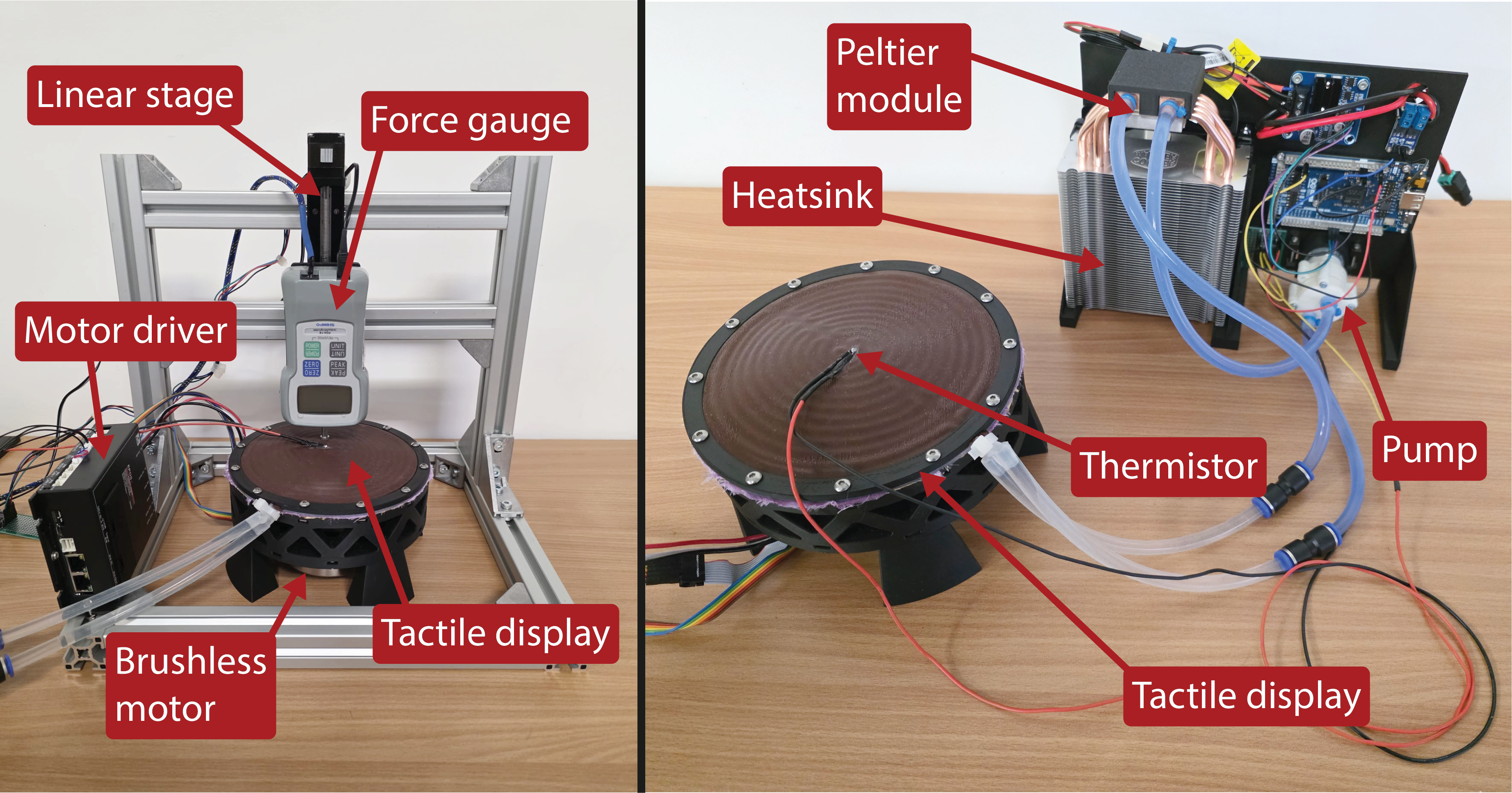}
\caption{Apparatus and setup used for initial contact stiffness (left) and temperature (right) measurement} \label{fig:setup}
\end{figure}

The softness of the device was assessed as the initial contact stiffness, since particle jamming systems are known to be actuated by the large contact force that would be incurred by measuring bulk stiffness \cite{Brown2022UsingJoystick}. Initial contact stiffness also has the benefit of being measured using very small forces, meaning that the effect of the silicone skin on hardness change will be reflected in the results. To generate and measure small contact forces, a digital force gauge (Shimpo FGV-1Z, capacity 5~N, sampling rate 20~Hz) with an 8~mm flat circular indenter was suspended above the tactile display from a linear motion stage. This was driven into the surface, up to a contact force of 4~N. This setup is shown in Figure \ref{fig:setup} (left). Stiffness was calculated as the average force-displacement gradient over the 0.5-1~N range, simulating initial contact with a finger.

Stiffness of the display was analysed as a function of motor torque (via closed-loop current control) from 0-2~Nm in increments of 0.2~Nm.

\subsection{Temperature}

The tactile display's achievable temperature range was assessed by passing water through the temperature controller at maximum heating power (open-loop control). Temperature change over time was recorded with an Amphenol 10~k$\Omega$ thermistor attached to the top surface. Measurements were taken every 0.25 seconds and the highest stable temperature recorded as a maximum. The minimum temperature was then tested by reversing the polarity of the Peltier module, maximising cooling power. The setup is shown in Figure \ref{fig:setup} (right). Surface uniformity at the minimum and maximum temperatures was measured with a thermal camera (Flir TG267, IR resolution 120 $\times$ 160), which was also used to investigate possible heat loss to the jamming fluid. This was repeated twice.

Closed-loop performance was assessed by commanding the display to reach and hold a moderate setpoint temperature. The time to reach this temperature was measured, as was the peak error once the setpoint was reached. All measurements were taken with the system described above.

\section{Results}


\subsection{Softness response}

\begin{figure}[h!]
\centering
\includegraphics[width=0.85\textwidth]{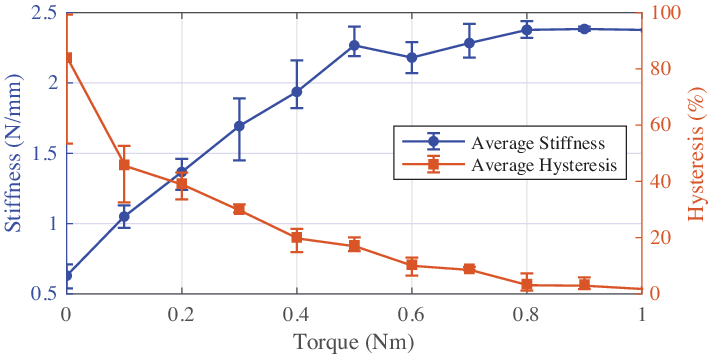}
\caption{Stiffness and hysteresis of the display under different motor torques. Error bars represent the highest and lowest values observed across repeated measurements.}
\label{fig:res-stiffness}
\end{figure}

The display exhibited an average minimum stiffness of 0.63~N/mm in its soft (unjammed) state. Average hysteresis in this state was 84\%. Stiffness then increased very linearly at an average \textit{stiffness sensitivity} of 0.00164~mm\textsuperscript{-2} up to 0.5~Nm (2.27~N/mm, hysteresis 10.4\%)~\footnote{Stiffness sensitivity is defined as the change of stiffness with respect to torque. Although unintuitive, mm\textsuperscript{-2} is the simplified form of $\frac{\mathrm{N/mm}}{\mathrm{Nmm}}$.}. Stiffness then remained broadly constant with increasing torque, increasing to only 2.38~N/mm at the maximum 1~Nm torque. Hysteresis, however, continued to decrease broadly exponentially, reaching 2.1\% at the maximum torque. Stiffness was generally consistent across repeated measurements, although with particularly low variance at high values (0.00016). Hysteresis deviation was consistently low, except in very soft states. The full stiffness results are shown in Figure \ref{fig:res-stiffness}.

\subsection{Thermal response}

\begin{figure}[h!]
\centering
\includegraphics[width=0.85\textwidth]{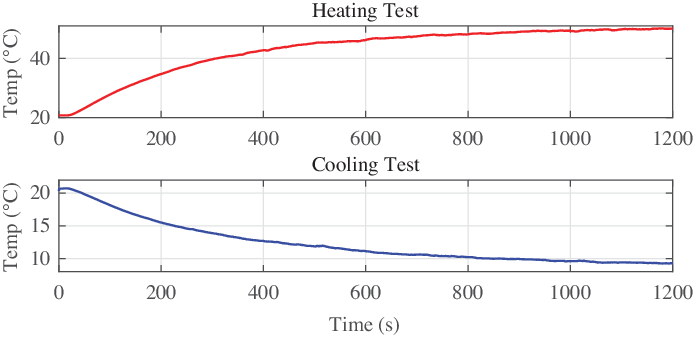}
\caption{Graphs of temperature change over time with heated and chilled water}
\label{fig:res-temp-range}
\end{figure}

When heated at full power, a maximum temperature of 50.2\degree~C was achieved in 20~minutes (thermal time constant $\tau$ = 290~seconds). There was a very small delay in the display temperature changing at the beginning of the test due to water circulating before being heated. When chilled, the display cooled to a minimum temperature of 9.3\degree~C ($\tau$ = 325.5~seconds). All thermal measurements were taken at an ambient temperature of 20.7\degree~C. These results are shown graphically in Figure \ref{fig:res-temp-range}. This test was conducted twice, with the display temperatures within 2\degree~C between tests. This is acceptable given variation in ambient conditions.

\begin{figure}[h!]
\centering
\includegraphics[width=0.8\textwidth]{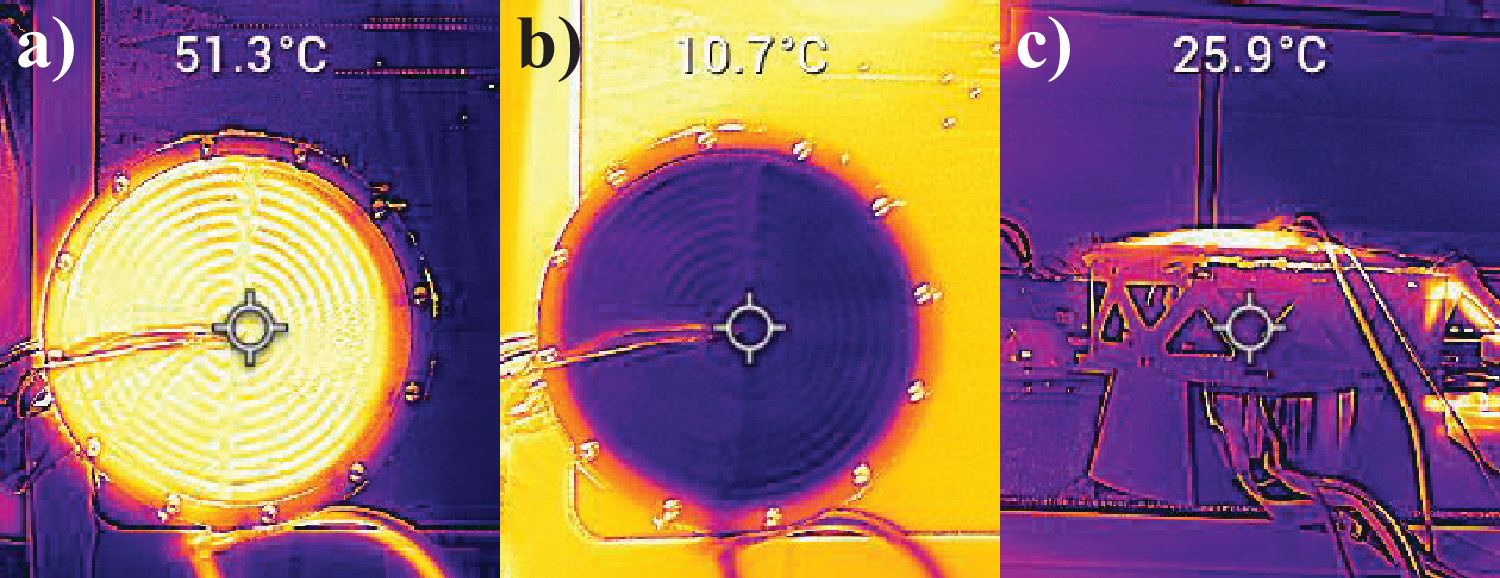}
\caption{Thermal images of the display at a) maximum and b) minimum temperature. Panel c) shows the side view during heating, demonstrating the limited heat loss into the large thermal mass of the particle fluid.}
\label{fig:res-temp-uniformity}
\end{figure}

The display temperature was generally uniform during heating, with a slight change in temperature between the two halves in the cold state. There was very little observable temperature difference between water channels (<0.3\degree~C). The display was slightly warmer (<2\degree~C) near the inlet, but then remained consistent. Due to the limited overall temperature change, uniformity was not assessed in the cold condition. Thermal images of the display close to its minimum and maximum temperatures are shown in Figure \ref{fig:res-temp-uniformity} a and b. Heat loss to the jamming fluid was minimal - the exterior of the pouch measured 26\degree~C at the maximum (>50\degree~C) surface temperature. Figure \ref{fig:res-temp-uniformity} shows a thermal image taken from the side of the device, demonstrating the concentration of heat in the touch surface.

\begin{figure}[h!]
\centering
\includegraphics[width=0.8\textwidth]{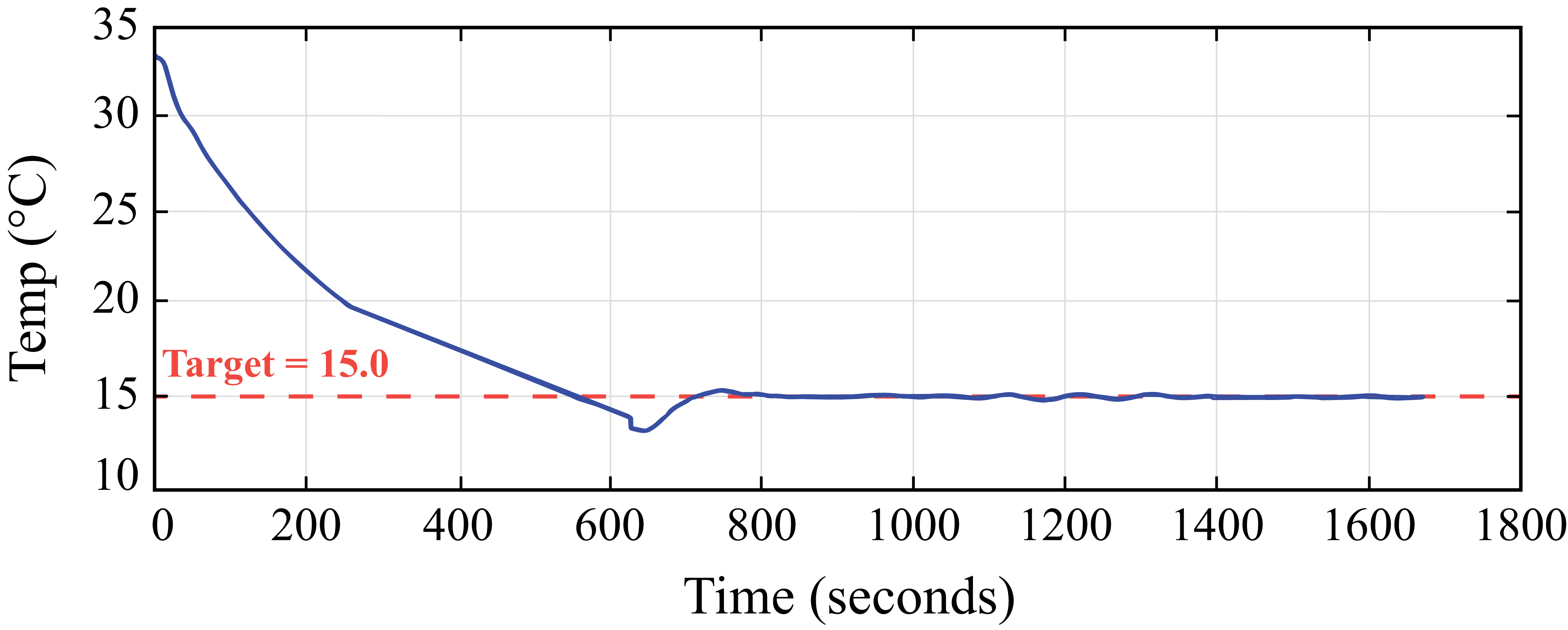}
\caption{Closed-loop performance of the display. Note that the monitoring system suffered a power fault at approx. 230 seconds, recovering at 620 seconds. The relevant performance indicators were not affected.}
\label{fig:res-temp-control}
\end{figure}

To evaluate closed-loop performance, the display was heated to approximately 33\degree~C, before being given a setpoint of 15\degree~C. It took 12 minutes to reach the setpoint value (an average temperature change of 1.5\degree~C per minute), after which the measured temperature remained within ±0.2\degree~C of the setpoint (Figure \ref{fig:res-temp-control}) for 15~minutes.

Throughout all tests, a 30~second latency was observed between sending a command to the display and observing an expected heating-cooling response.

\section{Discussion}

The prototype demonstrated a good range of achievable stiffness, despite initial contact stiffness being more susceptible to the low surface stiffness of the thermal skin. Stiffness was also highly controllable, with consistent responses to torque in the 0-1~Nm range. Hysteresis was initially high due to the elasticity of the thermal skin being able to build potential energy when depressed into the softer jamming substrate, but reduced quickly and was consistent throughout. This would have a noticeable effect on tactile sensation \cite{Ujitoko2022PerceptualIndentation}.

Thermal performance testing indicated some limitations. The display demonstrated good performance in heating, with surface temperature quickly matching human skin temperature. This has potential applications in domains such as medical simulation \cite{Genoud2026EnhancingInterface} and affective haptics \cite{Ackerley2014HumanCaress}. The rate of increase then slowed substantially. Whilst the maximum achievable temperature demonstrated a useful range, it was somewhat slow to reach higher temperatures. Possible reasons for this include the relatively large surface area of the touchpad causing energy loss. Smaller displays may perform better in the future, but are less useful for interactive tasks. Cooling was effective. The temperatures achievable within the 2-minute response time are potentially useful for simulating material properties \cite{Ho2018MaterialApplications}. Very low temperatures were achieved slowly. Future iterations of the technology will investigate the trade-off between display area and thermal response.

Surface uniformity was generally very good, with little evidence of the water losing energy as it flowed through the thermal skin. Temperature was also generally consistent above and between channels, indicating that the copper-silicone composite was effective in spreading thermal energy. The insulation provided between the thermal layer and the jamming substrate seemed to be effective based on external thermal photographs, indicating that the material design was capable of concentrating heat transfer to the display.

Preliminary evaluation of closed-loop control demonstrated sub-degree (typically ±0.2\degree~C) deviation. More in-depth testing is required to validate this performance robustly. Recovery from perturbation was not tested. The dynamic response was somewhat weak given the 30~second latency in the temperature change. This is likely due to poor thermal inertia in the Peltier module and fluid system, which itself has a relatively high thermal mass. Applications requiring faster hot-cold changes may benefit from dedicated hot and cold water supplies, with either hot or cold water being pumped depending on the target temperature. The system as presented is usable in perceptual research and medical simulation, which typically would not require rapid temperature changes.

\subsection{Future work}
Future work will seek to enhance the response rate by investigating the benefits offered by smaller display areas. The dynamic response will be improved by reducing the display size, and enhancing the liquid temperature control.

Scientific work will apply the new interface to address known challenges in the creation of multisensory experiences, as well as to investigate crossmodal correspondences involving both softness and temperature. Such experiences include medical simulation, where softness and temperature (heat or cold) are important indicators of infection, malignancy, and vascular disease, as well as affective haptic experiences that benefit from softness and warmth. Further investigations will explore the crossmodal perception of haptic interactions, such as the recognition of physical materials by their compliance and thermal conductivity, or the interaction of softness, temperature and weight - a variation of Weber's Illusion \cite{Goudge1918AIllusion}.

\section{Conclusion} 
This work demonstrates that particle jamming can be used as a foundation for multimodal tactile interfaces that can combine otherwise incompatible technologies to produce multiple distinct sensations simultaneously. This is evidenced through the combination of softness- and temperature-change in a desktop-scale haptic device. Future research will aim to improve dynamic performance, and use this capability to explore the perceptual characteristics and application areas of these crossmodal haptic interactions. 

\begin{credits}
\subsubsection{\ackname}
This work was supported by the Imperial College-MIT Global Seed Fund and EPSRC pro2 Network+, grant number EP/W020564/1.
J. Brown was supported by the Imperial College Digital Innovation Fund.

\subsubsection{\discintname}
The authors have no competing interests to declare.
\end{credits}
%
%
%
 \bibliographystyle{splncs04}
 \bibliography{joshua}

\end{document}